\def\year{2020}\relax
\documentclass[letterpaper]{article} 
\usepackage{aaai20}  
\usepackage{times}  
\usepackage{helvet} 
\usepackage{courier}  
\usepackage[hyphens]{url}  
\usepackage{graphicx} 
\usepackage{array}
\usepackage{subcaption}
\usepackage{xspace}

\newcommand{\BertSM}[0]{BERT$_{S+M}$\xspace}
\newcommand{\BertMaskS}[0]{BERT$_{MaskS}$\xspace}
\newcommand{\BertMaskSM}[0]{BERT$_{MaskS+M}$\xspace}
\newcommand{\BAMaskS}[0]{BA$_{MaskS}$\xspace}
\newcommand{\BAMaskSM}[0]{BA$_{MaskS+M}$\xspace}

\newcommand{\citet}[1]{\citeauthor{#1}~\shortcite{#1}}
\newcommand{\citep}{\cite}
\newcommand{\citealp}[1]{\citeauthor{#1}~\citeyear{#1}}

\title{Corpus Wide Argument Mining - a Working Solution}

\author{Liat Ein-Dor, Eyal Shnarch, Lena Dankin, Alon Halfon, Benjamin Sznajder,\\
\bf \Large Ariel Gera, Carlos Alzate, Martin Gleize, Leshem Choshen, Yufang Hou, \\
\bf \Large Yonatan Bilu, Ranit Aharonov and Noam Slonim\\
IBM Research AI\\
\{liate,eyals,lenad,alonhal,benjams,arielge,leshem.choshen,yonatanb,ranita,noams\}@il.ibm.com\\ \{carlos.alzate,martin.gleize,yhou\}@ie.ibm.com
}

\begin{document}
\maketitle
\begin{abstract}
 One of the main tasks in argument mining is the retrieval of argumentative content pertaining to a given topic. Most previous work addressed this task by retrieving a relatively small number of relevant documents as the initial source for such content. This line of research yielded moderate success, which is of limited use in a real-world system. Furthermore, for such a system to yield a comprehensive set of relevant arguments, over a wide range of topics, it requires leveraging a large and diverse corpus in an appropriate manner. 
 Here we present a first end-to-end high-precision, corpus-wide argument mining system. This is made possible by combining sentence-level queries over an appropriate
 indexing of a very large corpus of newspaper articles, with an iterative annotation scheme. This scheme addresses the inherent label bias in the data and pinpoints the regions of the sample space whose manual labeling is required to obtain high-precision among top-ranked candidates.
\end{abstract}

\section{Introduction}
Starting with the seminal work of \citet{MochalesPalauRaquel2009Amtd}, argument mining has mainly focused on the following tasks - identifying argumentative text segments within a given document; labeling these text segments according to the type of argument and its stance; and elucidating the discourse relations among the detected arguments. Typically, the considered documents were argumentative in nature, taken from a well defined domain, such as legal documents or student essays. More recently, some attention had been given to the corresponding retrieval task - given a controversial topic, retrieve arguments with a clear stance towards this topic. This is usually done by first retrieving -- manually or automatically -- documents relevant to the topic, and then using argument mining techniques to identify relevant argumentative segments therein. This documents-based approach was originally explored over Wikipedia \cite{levy-etal-2014-context,rinott-etal-2015-show}, 
and more recently over the entire Web
\cite{stab2018cross}. 
This approach is most suitable for topics of much controversy, where one can find documents directly addressing the debate, in which relevant argumentative text segments are abundant.

For an argument retrieval system to be of practical use requires: (1) 
high precision, and (2) wide coverage. The former is important because in order to make a case one typically employs several arguments, all of which need to be relevant and persuasive. For example, if one aims to retrieve $3$ such arguments, then even a precision of 0.9 might not be enough, as it implies a $0.27$ probability of getting at least one of the arguments wrong. While previous work made great advancements in argument retrieval, such 
high precision was far from being attained.
The second requirement is what sets a practical system from a proof-of-concept. Indeed, previous work mostly focused on a small set of topics and a corresponding moderate-sized corpus. Here we achieve wide coverage by demonstrating how to perform corpus--wide argument mining over massive corpora.

Our starting point is the approach of \citet{Levy2017UnsupervisedCC}, which
was the first to depart 
from the document-based dogma, 
aiming 
to detect topic-relevant arguments by indexing the \textit{sentences}
of a corpus, and retrieving them directly. This Sentence-Level (SL) approach can potentially detect a much 
larger set of argumentative texts taken from a wider set of articles, including articles not focused on the topic directly, and which are probably overlooked by the document-level approach. 

On the other hand, a main challenge emerging from the SL approach is the skewness of the data. When such an approach is applied naively, only a very small fraction of the data are positive examples (relevant arguments). Such an imbalanced dataset is known to be the bane of many supervised machine-learning algorithms. Indeed, this problem is also apparent in the document-level approach, in cases where it is common for the topic to be discussed also in non-argumentative contexts, but is more acute with the SL approach. Previous work either focused on topics for which argumentative content is abundant, or used weak-supervision to try and mitigate the problem.

Our approach is different from previous SL works in three important aspects: (1) We demonstrate 
for the first time 
how to leverage supervised--learning techniques for this task, achieving considerably stronger results than earlier works, which adopted the weak--supervision paradigm; 
(2) We develop an efficient labeling methodology, reminiscent to 
active learning, which, alongside query-based filtering of the corpus, addresses the data imbalance problem; 
(3) we report results over a massive newspaper corpus, containing close to $10$ billion sentences, while earlier SL works reported results only for Wikipedia, which is $\sim50$ times smaller. We note that the queries used in (2) are similar in essence to the rules used for weak supervision by \citet{Levy2017UnsupervisedCC}.

Our approach consists of two steps. First, dedicated queries are applied to retrieve an initial set of potentially relevant sentences. Second, a classifier determines which of the retrieved sentences are both argumentative and relevant. To train such a classifier, one needs to construct a corresponding labeled dataset, which ideally is balanced between the positive and negative examples. However, the 
fraction of argumentative sentences 
among those retrieved is typically rather small.
To overcome this, we propose 
an iterative {\it retrospective-labeling} approach, of labeling the top predictions of the system, and then training on the obtained labeled data. To bootstrap the process, we start with a modest-sized set of labeled data - for example, one obtained via the document-based methods.

The resulting approach is precision-oriented. Clearly, arguments within sentences that do not satisfy the queries we start with, will be missed by such a system. Nonetheless, the experimental results herein suggest that when this approach is coupled with a very large corpus 
it is able to retrieve, with high precision, a large and diverse set of arguments for a diverse set of topics.

Importantly, any "real world" retrieval system needs to be efficient - it is unfeasible to classify each sentence in the corpus for whether or not it is a relevant argument. Thus, some filtering method must be applied, be it document-retrieval, which limits the scope to the sentences of the retrieved documents, or, as we do here, SL, which limits the scope to sentences matching the queries.

Arguments can be varied and complex. Perhaps the simplest ones, which also serve as building blocks for more complex arguments, are {\em Claims} and {\em Evidence} \cite{toulmin}. Thus, the complexity of the problem we examine may vary according to the the types of arguments sought; asking for a very specific argument type is a more difficult problem, since these are harder to come by and because the problem is compounded by the need to discern their type. Importantly, in the document-level approach, it is not clear how to aim for documents that are rich with a specific type of arguments. Our SL approach is more flexible in this sense, since by properly designing the queries, one can explicitly search for arguments of a desired type. 

To demonstrate the applicability of our approach, 
we focus on the retrieval of {\em Evidence}, specifically those \citet{rinott-etal-2015-show} denoted as {\em Expert Evidence} and {\em Study Evidence}. Thus, our system can be perceived as a tool that facilitates critical evaluation of a given topic by providing 
evidence which support or contest the common claims and beliefs surrounding the topic. In this manner it may be useful to alleviate some of the concerns around fake news.
Nonetheless, the methodology presented here is applicable to other argument types as well. 

In summary, the main contribution of this work is in presenting a high--precision wide--coverage 
argument retrieval system, 
validated over a massive real-world corpus, for a diverse set of topics. Importantly, this is achieved by a new labeling paradigm 
that yields
a large and balanced annotated dataset over this corpus, and by
combining SL queries with a highly effective supervised-learning classifier. 

\section{Related Work}
Starting with the work of \citet{MochalesPalauRaquel2009Amtd}, who looked at legal documents, argument mining has initially focused on documents from specific domains, which tend to be argumentative by nature. In addition to legal documents \cite{Moens:2007:ADA:1276318.1276362,Wyner:2010:ATM:2167945.2167950,Grabmair:2015:ILE:2746090.2746096}, examples also include student essays \cite{DBLP:journals/corr/abs-1802-05758,persing2016end}, user comments on proposed regulations \cite{park-cardie-2014-identifying} and newspaper articles \cite{Feng:2011:CAS:2002472.2002597}. 
As with other fields in NLP, deep learning has proven to be a powerful tool for such tasks, be it direct argument mining \cite{eger-etal-2017-neural}, predicting argument persuasiveness \cite{habernal-gurevych-2016-argument,gleize2019you}, 
detecting context dependent claims and evidence \cite{DBLP:journals/corr/LahaR16}, or inferring relations among arguments \cite{cocarascu-toni-2017-identifying,persing2016end}. 
See, for example \cite{cabrio2018five,lawrence2019online} for recent reviews of the field.

Corpus-wide argument mining was originally suggested by \citet{levy-etal-2014-context}. They considered 32 controversial topics, manually identified 326 relevant Wikipedia articles, and labeled some 50000 sentences appearing in these articles for whether or not they contain a claim with a clear stance towards the topic. Similarly, \citet{rinott-etal-2015-show} aimed to detect evidence for 58 different topics,
over 547 manually-collected articles, and \citet{stab2018cross} did so for 8 controversial topics, while considering the top 50 Google-retrieved web pages for each topic.

These methods have achieved moderate success. For example, \citet{levy-etal-2014-context} report that among their top 50 predictions, on average, precision is 0.19 and recall is 0.4 (a random baseline achieves precision of 0.02). In \citet{stab2018cross}, where 44\% of the sentences are positive examples, and all sentences are classified, the best F1-score is 0.66 (an "all-yes" baseline has an F1-score of 0.61). 

We envision a "real world" argument retrieval system as one which retrieves several arguments, that are typically all relevant to the topic. Hence, such a system requires 
high precision, well beyond the aforementioned accuracy-oriented systems. Moreover, it should work well on a wide variety of topics, many of which are probably not associated with such argument-abundant articles as in the data of \citet{stab2018cross}. Both of these goals are achieved in this work.

Other work on argument retrieval includes \citet{wachsmuth-etal-2017-building} and \citet{al-khatib-etal-2016-cross}, who also consider cross-topic argument mining, but do so over a dataset of arguments rather than a heterogeneous corpus where only a small fraction of the text is argumentative. Similar to our work here, \citet{shnarch-etal-2018-will} also aim to classify sentence candidates for whether or not they are evidence, but their emphasis is on a semi-supervised method, applied to Wikipedia, without addressing the sentence-retrieval task. 

The iterative retrospective labelling technique suggested here may seem similar to the semi-supervised approach of Label Propagation \cite{zhu2002learning}. In Label Propagation, unlabeled data is pseudo-labeled automatically according to the labels of similar data, and similarity is computed automatically. By contrast, we do not attempt to define similarity between arguments, and unlabeled data of interest is manually labeled. 

Another related labelling approach is Active Learning \cite{cohn1996active}, an umbrella term covering a variety of algorithms which aim to select informative instances for labelling, in order to learn a suitable classifier. 
Previous studies have indicated that the performance of active learning is easily disrupted by an imbalanced data distribution \cite{zhu2007active,bloodgood2009taking}. 
Moreover, in most active learning literature, accuracy
is chosen as the evaluation metric, while in our case, the quantity of interest is precision. A recent work \cite{wang2018uncertainty} suggested a method for optimizing expected average precision in active learning, but did not deal with the problem of skewed class distribution. To the best of our knowledge, our iterative retrospective labelling technique is the first precision-oriented active learning strategy for coping with the class-imbalance problem. Moreover, while \citet{wang2018uncertainty} are interested in average precision, which is affected by the entire set of examples, in our task it is the top-ranked predictions which are of interest.

\section{Definitions}\label{sec:def}
Following the nomenclature of competitive debate and previous papers (e.g. \citealp{levy-etal-2014-context}), we define a {\em Motion} as a high-level claim implying some clearly positive or negative stance towards a (debate's) {\em topic}, and, optionally some policy or {\em action} that should be taken as a result. For example, a motion can be {\em We should ban the sale of violent video games} or {\em Capitalism brings more harm than good}. In the first case, the topic is {\em violent video games}, and the proposed action is {\em ban}. In the second case, the topic is {\em Capitalism}, and no action is proposed. In this work, the topic of motions will always be a reference to a Wikipedia article.

In the context of a motion, we define 
{\em Evidence} as a single sentence that clearly supports or contests the motion, yet is not merely a belief or a claim. Rather, it provides an indication whether a belief or a claim is true.

Specifically, we are interested in two types of Evidence: {\em Study Evidence}, which presents 
a quantitative analysis of data, and {\em Expert Evidence} which presents 
testimony by a relevant 
expert or authority \cite{rinott-etal-2015-show}.

For example, the sentence {\em The research clearly suggests that, among other risk factors, exposure to violent video games can lead to aggression and other potentially harmful effects} is {\em Study Evidence} supporting the motion, and specifically
supports the claim that {\em violent video games can lead to aggression}.

\section{Data Acquisition}
In supervised learning, class imbalance in training data distribution often causes
learning algorithms to perform poorly on the minority class.
This issue has been well addressed in the machine learning literature \cite{chawla2004special,garcia2010theoretical,van2007experimental}. The corpus we consider in this work is very large, containing some 400 million articles, in which, for nearly any motion, positive examples are few and far between. This poses the question of how to obtain training data which is relatively balanced, while not wasting much of the labeling effort on sentences which would later be discarded.

Our approach is as follows. Start with a small set of manually collected and labeled sentences - which can be obtained by manually querying the corpus, or by taking data from a smaller corpus, which is enriched with positive examples. Next, train a simple classifier, 
such as logistic regression, 
which can generalize well based on this dataset, 
and use it to predict the label of many relevant 
sentences in the very large corpus 
(e.g., sentences that satisfy specific queries).
Labeling
the top predictions of the classifier, one obtains a larger dataset.
Note that in this second phase the set of motions can be expanded well beyond the original one, further increasing the size of the dataset.

Labeling the top predictions has two advantages. First, since they are enriched with true-positive examples, this helps lessen the label imbalance in the dataset. Second, it brings forth the "hard" negative examples; those which are likely to be the main obstacle for obtaining what we seek for a "real world" retrieval system: a high precision among the top-ranked candidates. By labeling these "hard" examples the classifier is improved in the relevant regions of the sample space.

Once the dataset is large enough, and not too skewed, one can train a more powerful classifier, such as a neural net. The dataset and classifier can then be improved in a similar manner, by iteratively labeling the top predictions of the new classifier, and presumably increasing the fraction of positive examples in each iteration.

Specifically, our starting point is a corpus of some 400 million newspaper and journal articles provided by LexisNexis (which we denote as VLC - Very Large Corpus),
 and the logistic-regression classifier constructed by
\citet{rinott-etal-2015-show}, which is based on a 
relatively 
small set of manually labeled Wikipedia sentences. We selected 192 and 47 motions for our train and development sets respectively, and for each motion retrieved sentences in the corpus using queries similar to those in \citet{Levy2017UnsupervisedCC} (see Section \ref{sec:arch}). 
The retrieved sentences were then ranked using the logistic-regression 
classifier, and the top 40 sentences for expert and study evidence were manually annotated using crowd-sourcing\footnote{Using the Figure Eight platform, www.figure-eight.com.}. 
Annotation was binary - whether or not the sentence is an Evidence of the desired type - and the gold label was determined by majority.

The network of \citet{shnarch-etal-2018-will} was then trained on this dataset, and again the top 40 predictions for each motion and each evidence type were manually annotated. This was iterated until we ended up with a dataset of $198,457$ manually labeled sentences, of which 33.5\% are positive examples. We denote this dataset as the Very Large Dataset (VLD).

During the labeling process, each sentence-motion pair was annotated by 10 Figure-Eight annotators. Inter-annotator agreement (over multiple tasks) was computed via Cohen's Kappa for all pairs of annotators\footnote{Whenever agreement was computed, only annotators with at least $50$ items in common were considered.}. 
Then, for each annotator, the average agreement with other annotators - weighted by the number of common items - was computed. Annotators with low average agreement (Cohen's Kappa $< 0.3$) were discarded. If the number of labels for a sentence-motion pair was below 7, we kept labeling to reach 7 trusted annotations. This yielded an overall average Cohen's Kappa of 0.47 -- computed over all pairs of remaining annotators, and weighted by the number of common items. Labels of sentence-motion pairs were determined by majority; ties were taken as negative labels.

Note that a Cohen's Kappa of 0.47 is on par with many previous reports, focusing on \textit{type-dependent} argumentation datasets \cite{stab2018cross,stab2017parsing,boltuzic-snajder-2014-back}, reflecting the challenging nature of the task compared to \textit{type-independent} annotation \cite{levy-etal-2014-context,rinott-etal-2015-show,shnarch-etal-2018-will}. Finally, as indicated by the very high precision of the resulting models, the dataset evidently 
provides a clear signal for evidence detection. 

In addition to the dataset above, a matching dataset over Wikipedia was constructed. For the same set of training and development motions as in the VLD, the queries and baseline network described in section \ref{sec:arch} were used to retrieve and rank Wikipedia sentences. The top 20 predictions for each motion were then manually annotated. This yielded $29,429$ 
labeled sentences (of which 23\% are positive examples), which we denote as the Wikipedia Dataset (Wiki) 
and is 
available at \url{http://ibm.biz/debater-datasets}.

\section{System Architecture}\label{sec:arch}
Sentence retrieval is done using queries that specify which terms should appear and in what order, with some gaps allowed. These terms include the {\em topic}, the {\em action}, named entities such as numbers, persons and organizations, lexicon terms indicating sentiment, or of particular relevance to the type of evidence sought - Expert or Study, and certain connectors which are indicative of evidence. Importantly, all queries require the {\em topic} to appear in the sentence, and so retrieved sentences always contain a corresponding term. 

For example, a query oriented towards retrieving Evidence of type Study is defined by requiring that the following elements appear in a sentence in order (but not necessarily adjacent; other words may appear in between): 
(1) a term from a study related lexicon; (2) the conjunction term {\em That} \cite{LevyBGAS18}; (3) the {\em topic}; (4) a term from a sentiment lexicon. Accordingly, for the topic \textit{gambling}, one of the sentences retrieved by the query is: \textit{The University of Glasgow and Healthy Stadia \textbf{research} warns \textbf{that} \textbf{gambling} is a public health issue with potential for \textbf{harm}}, where the words in bold are those that match the required terms (in order).

A full description of the queries and their associated lexicons is available along with the Wikipedia Dataset files.

Queries are arranged in a cascade, and a total of up to $12,000$ sentences is sought for each evidence type. Queries that are more likely to yield evidence are run first (see Supplementary Materials), and the cascade halts once the $12,000$ 
limit has been reached. 

Allowing such queries required indexing the VLC not only for word strings, but also for their relevant semantic roles, such as being a named entity or appearing in one of the lexicons. Moreover, since in our framework the {\em topic} is a Wikipedia title by definition (see section \ref{sec:def}), 
all sentences are also "wikified", that is, terms are linked to their underlying Wikipedia titles when possible.
To allow processing billions of sentences in a timely manner, we use a simple rule-based Wikification method which mainly relies on Wikipedia redirects \cite{shnayderman2019fast}.

\begin{figure}[ht!]
\centering
\includegraphics[width=.9\columnwidth]{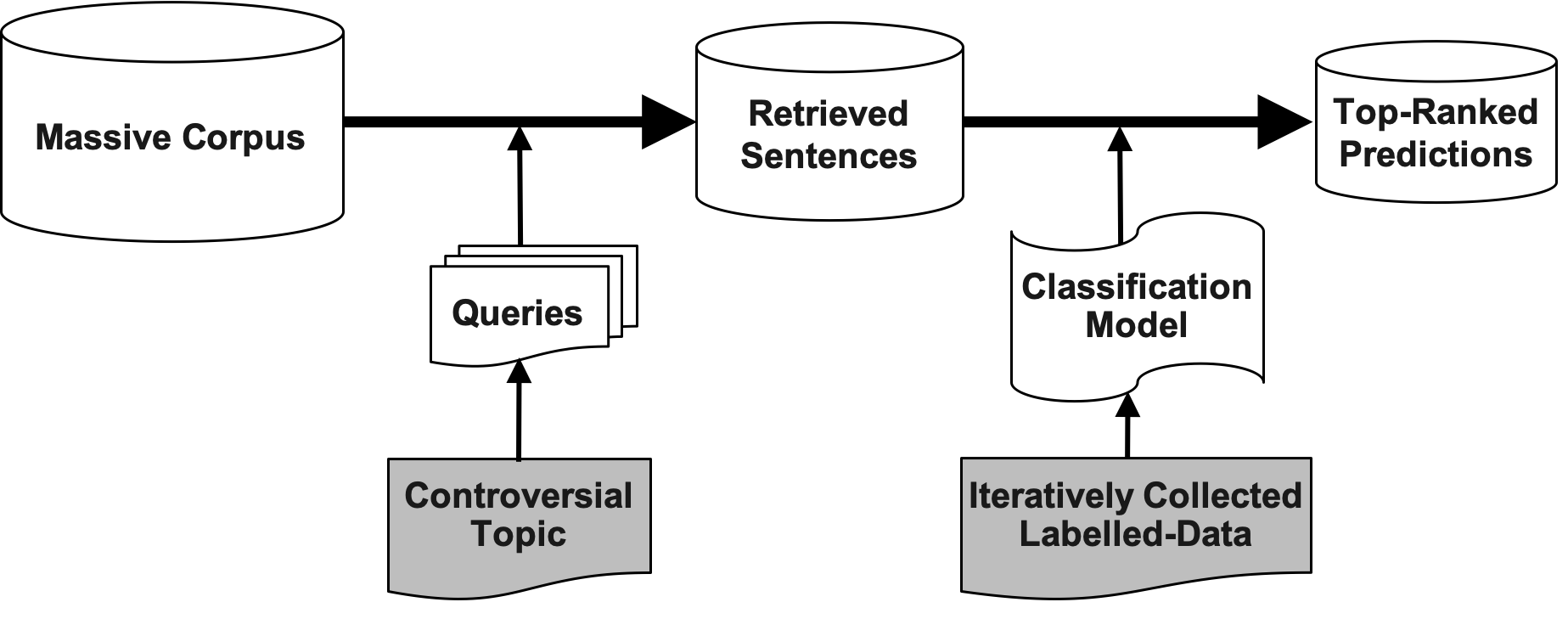}
\caption{Overview of the end-to-end evidence retrieval system. 
\label{fig:arch}}
\end{figure}

The output of the retrieval stage is the union of sentences from both the {\em Expert Evidence} queries and the {\em Study Evidence} queries, with duplicate sentences removed. 
Figure \ref{fig:arch} summarizes the suggested pipeline for evidence retrieval.

Whereas recent work on evidence retrieval \cite{stab2018cross,shnarch-etal-2018-will} was 
limited to considering a small set of documents and correspondingly focused 
on the {\it classification\/} 
task, here we describe a full end-to-end retrieval system, and hence focus 
on {\it ranking\/} the 
retrieved sentences. 
Specifically, 
retrieved sentences are fed to a classifier which computes a score indicating the confidence that the sentence is an evidence to the motion, and ranked according to this score.

\section{Experiments}

\label{experiment}
To evaluate the suggested end-to-end framework we trained several neural net variants on the VLD, and assessed their performance on four benchmarks. 

The network variants differ in the input they receive and in the underlying architecture.  We consider the following input variants: 
\begin{itemize}
    \item S+M - sentence and motion
    \item MaskS+M - masked sentence and motion
    \item MaskS - masked sentence
\end{itemize} 
Masking is done by replacing the {\em topic} in a sentence with a special token (recall that all retrieved sentences contain the {\em topic}). The purpose of masking is threefold: 
(1) it marks the topic within the sentence; (2) it introduces a uniform representation for the different forms of the same topic; (3) it introduces a uniform representation for the different possible topics.

Masking was also used by \citet{shnarch-etal-2018-will}, but they did not consider adding the motion as input. Hence, they in fact framed the task as deciding whether or not a sentence is evidence for some implicit motion. By making the motion explicit, the network's training is ostensibly more aligned with what it is actually meant to do - determine whether or not a sentence is evidence for a {\em given} motion.

We examine two families of network architectures:
\begin{itemize}
    \item BA - the network described in \citet{shnarch-etal-2018-will}. A bi-directional LSTM \cite{graves2005framewise} with an additional self attention layer \cite{yang2016hierarchical}, followed by a fully connected layer.
    \item BERT - a network based on BERT \cite{devlin2018bert}, where the pre-trained model is fine-tuned using our labeled dataset. 
\end{itemize}
The motivation for choosing BA for our experiments is that it was successfully used by \citet{shnarch-etal-2018-will} for sentence classification in the context of evidence detection. We aimed to determine the effect of training this network on the much larger dataset that we were able to collect with the methods described above, and to see how the different input variants affect its performance. 
We chose BERT since it presents state-of-the-art results in a wide variety of NLP tasks.
Notice that BA receives a single text as input. In order to train a BA-based network which also receives the motion as input (i.e., the S+M and MaskS+M variants), we made the following modifications to the embedding layer of BA.
We embed both the sentence and the topic using the same BA embedding layer (BiLSTM+attention). We then obtain the joint embedding by concatenating the subtraction and the pointwise product of the two resultant embeddings, and add a fully connected layer on top of the joint embedding to obtain the output. All BA-based models were trained with the cross-entropy loss function.

The four benchmarks used for evaluation are as follows. 
The first and second benchmarks were obtained by running an end-to-end system on large corpora - one being the VLC, and the other Wikipedia (2018 dump) - for 100 novel motions. Top predictions were then manually annotated to directly assess ranking, and precision was computed as a function of the number of predictions made (annotation is done in the same way as for the training data, and over the top predictions of all network variants). For completeness, we also trained all networks on the Wiki dataset, and assessed their end-to-end precision over the Wikipedia benchmark. 

Importantly, the aim of the evaluation is to simulate the end-to-end performance on unseen motions, focusing on the precision of top predictions. Hence, using a fixed test data is problematic, since it does not reflect the predictions of the considered model, but a somewhat arbitrary set accumulated during the iterative labeled data collection.

Note that results are comparable only when considering the same benchmark, but not between benchmarks. The VLC, containing some 100 times more documents, is expected to have far more relevant evidence than Wikipedia. So, for example, when computing the precision over the top $k$ predictions, Wikipedia might simply not contain $k$ relevant sentences, and a precision of, say, $1.0$ would not be attainable. However, for a given benchmark, since all networks rank the same set of sentences, precision results are comparable. Note also that since the VLC is so big, it is not uncommon to retrieve paraphrases of essentially the same evidence - much more so than in Wikipedia. To mitigate this problem, when ranking VLC sentences, those with high similarity\footnote{Word overlap of at least 0.8 w.r.t. the shorter sentence; excluding stop-words and the topic.} to higher-ranking ones were removed. This filtering resulted in the removal of $~10\%$ of the retrieved sentences.

The third and fourth benchmarks compare our system to previous works - \citet{shnarch-etal-2018-will} and \citet{stab2018cross} - in which the sentences are given, and the goal is to classify them. We denote these benchmarks as {\em BlendNet} and {\em UKP-TUDA}, respectively. We evaluate our networks on the benchmarks in the same way as in the original works - computing accuracy on {\em BlendNet}, and precision-recall on {\em UKP-TUDA}. To do this, networks' scores are converted to binary labels by taking a threshold of $0.5$ (all variants were trained with loss functions w.r.t. binary labels). 

Since the original works did not filter out similar sentences, 
we did not do so for these benchmarks. Additionally, as some of the motions in the test set of \citet{shnarch-etal-2018-will} also appear in the VLD training and dev sets, models were retrained for the purpose of evaluation on this benchmark, by excluding motion-sentence pairs corresponding to overlapping motions from the train and development sets.

\section{Results}\label{sec:results}

\subsection{End-to-end System}
We evaluate the end-to-end performance of the different models using two benchmarks, VLD and Wiki on the $100$ test motions.
\subsubsection{VLD}
\begin{figure}[ht!]
\centering
\includegraphics[width=\columnwidth]{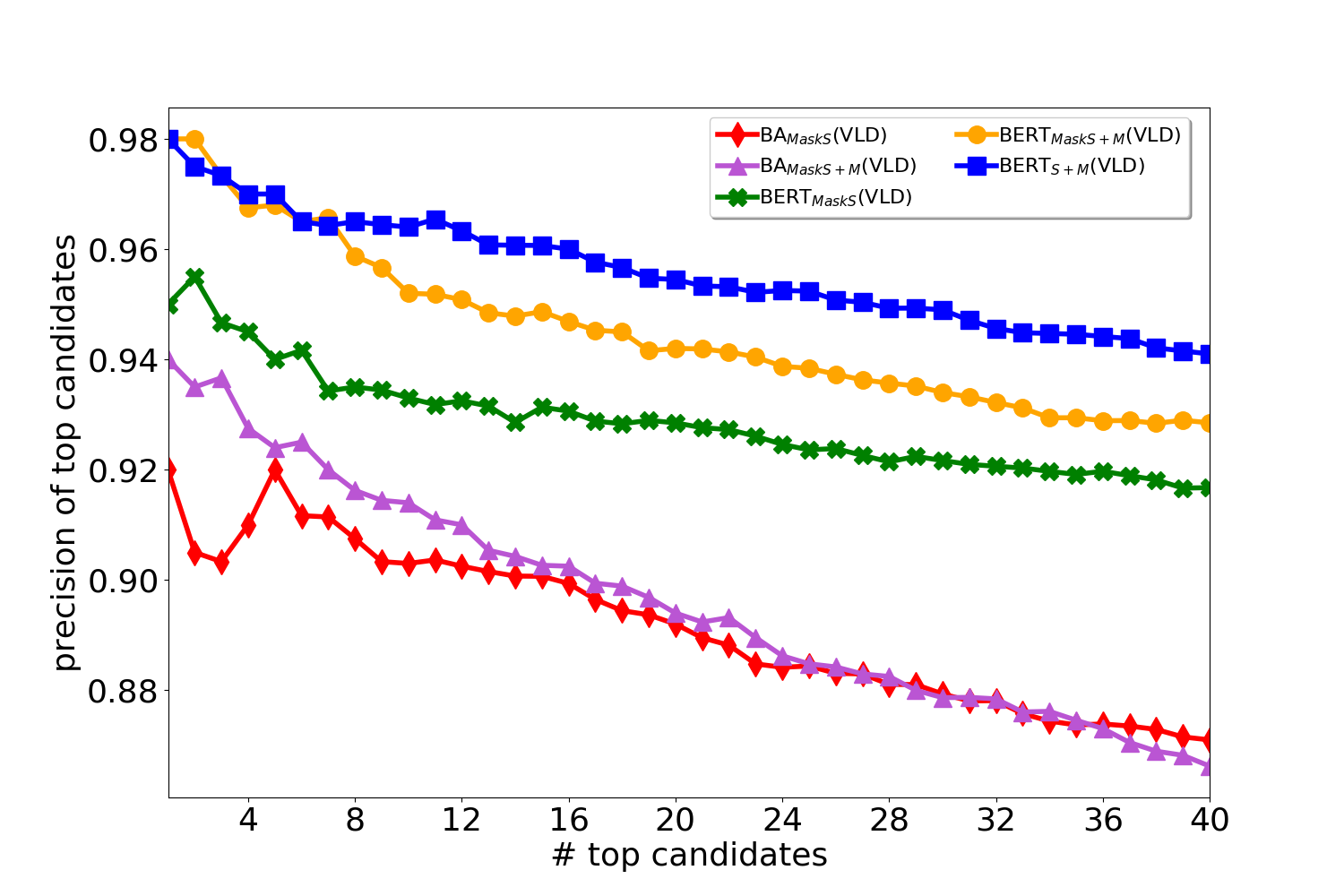}
\caption{VLD: Precision of top $k$ candidates vs. $k$ for different evidence
detection models. 
\label{fig:VLD_test}}
\end{figure}

Figure \ref{fig:VLD_test} presents the average precision of different evidence detection models as a function of the number of top candidates per motion. Precision is high for all models, with over $90\%$ precision for the top $20$ candidates, 
and a remarkably high precision of $95\%$ for the best model over the top $40$ candidates. 
This compares to an estimated positive prior of $0.3$ among the sentences retrieved from the queries\footnote{The estimate is based on labelling $10$ random sentences for each of the $100$ test motions.}.
The best performing model is \BertSM, with precision values that clearly surpass those obtained by the \BAMaskS model used by \citet{shnarch-etal-2018-will}. Notice that the \BAMaskS model receives only the masked sentence as input, while \BertSM receives the unmasked sentence and the motion. 
As expected, both BA and BERT benefit from the addition of the motion text to the input (see Figure \ref{fig:VLD_test}).

Among the BA based models, the best results are achieved when the input is the masked sentence and the motion text (\BAMaskSM). As mentioned in Section \ref{experiment}, the advantage of masking can be attributed to the extra information it provides to the network. First, it conveys to 
the network where the topic is located within the sentence. Second, the topic may appear in multiple textual forms. For example, the topic “capital punishment” may appear in the text as “execution”, “death penalty”, etc. Replacing Wikified instances of the topic with a single mask saves the need to learn the equivalence between the different forms. Finally, the usage of the same masking token for all topics can help the network learn general, motion-independent features of evidence texts.

In the case of BERT, the masking hinders the network performance. Models that are fed with masked sentences are inferior to \BertSM, which receives an unmasked sentence and a motion. A possible explanation for the negative effect of masking on BERT is that its strong underlying language model enables it to deduce the aforementioned information provided by masking from the original unmasked 
text. 
Consequently, the information that is lost in the masking process, i.e. the masked tokens, leads to inferior results.

\begin{figure}[ht!]
\centering
\includegraphics[width=.9\columnwidth]{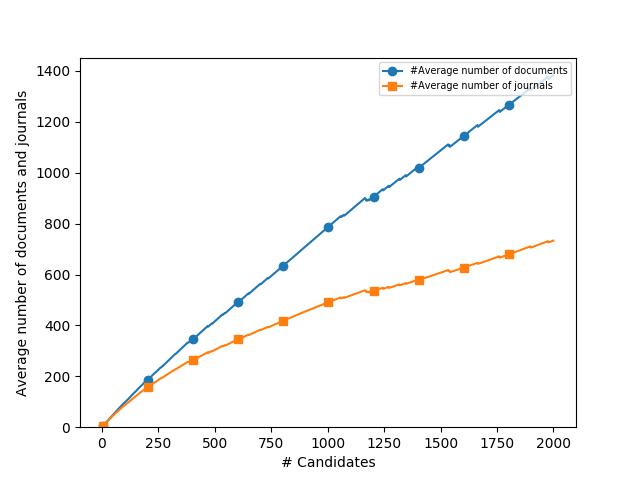}
\caption{Average number of documents and journals from which the top $k$ candidates originate, as a function of $k$. Results are based on top candidates of \BertSM(VLD).
\label{fig:diversity}}
\end{figure}

It is also interesting to evaluate to what extent the direct sentence-retrieval approach differs from one based on document-retrieval. In other words, do high-ranking sentences come from a wide range of documents and journals, or is this approach {\em de-facto} similar to the document-based one, with high-ranking sentences originating from 
only a small number of documents? 
As shown in Figure \ref{fig:diversity}, the former is true. For example, the top $20$ and top $40$ ranked candidates per motion from the \BAMaskS model originate from an average of 18.03 and 36.07 different documents, respectively.

\begin{figure}
\centering
\begin{subfigure}[b]{0.8\columnwidth}
\centering
\includegraphics[width=\columnwidth]{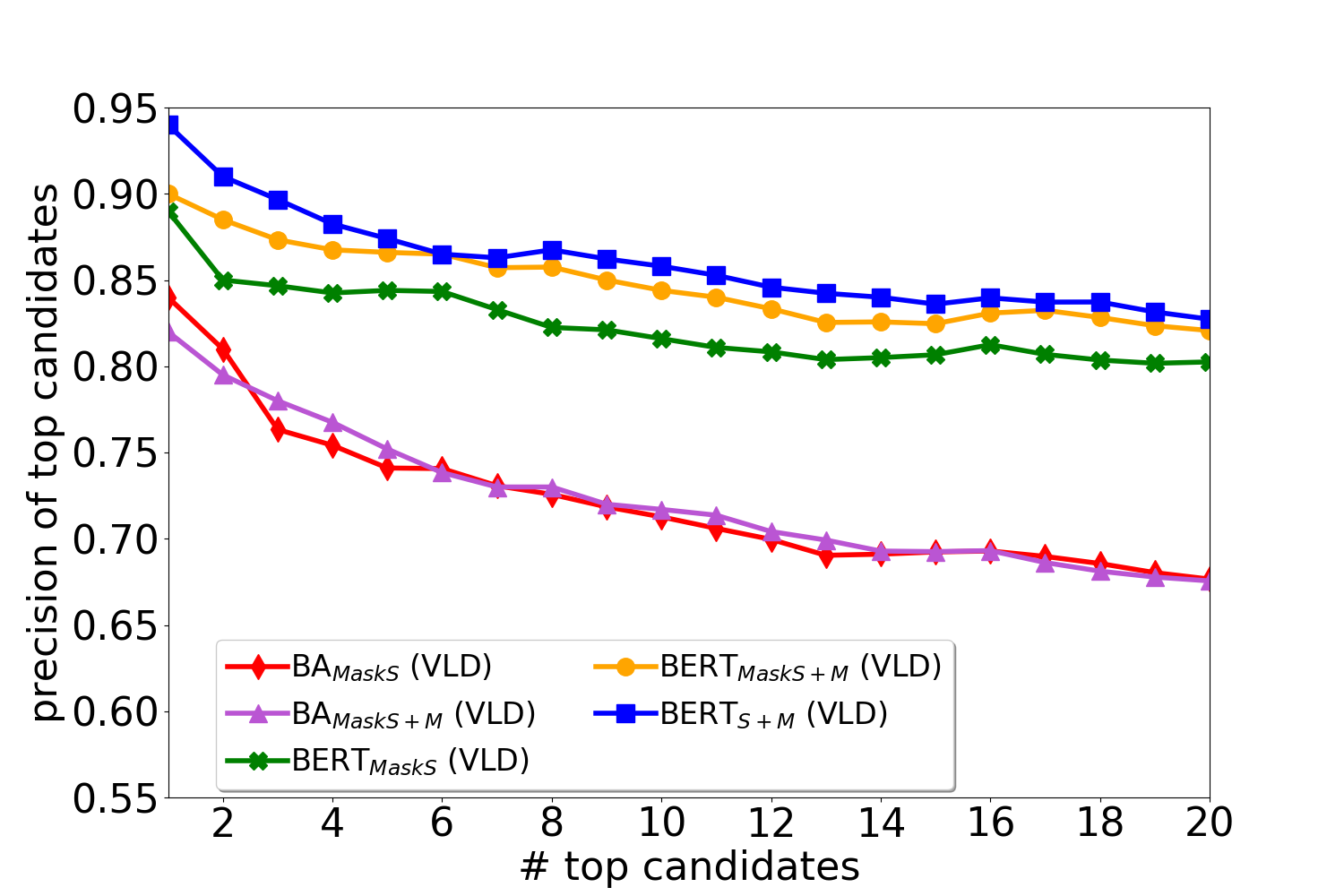}
\caption{Trained on VLD}
\end{subfigure}
\begin{subfigure}[b]{0.8\columnwidth}
\centering
\includegraphics[width=\columnwidth]{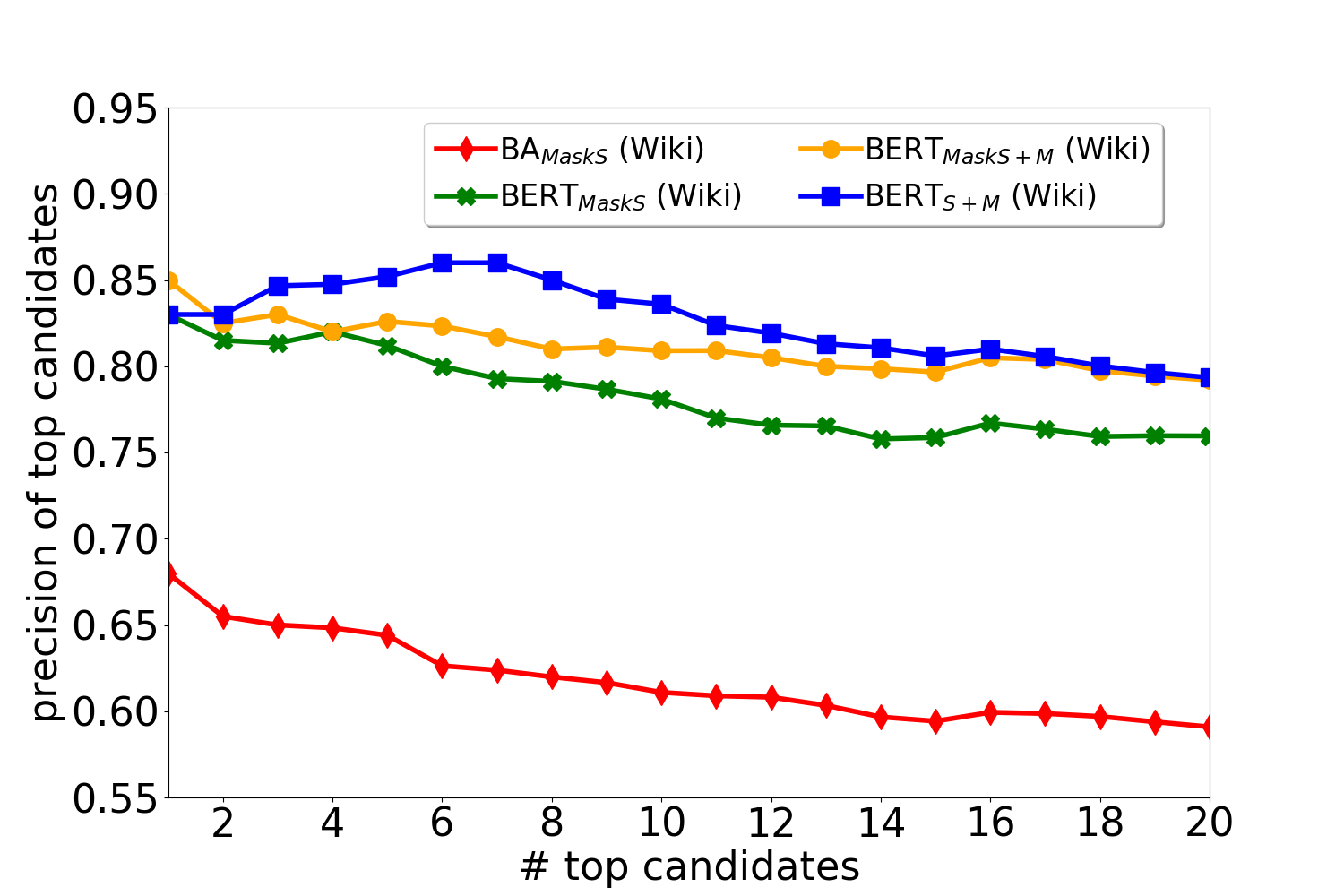}
\caption{Trained on Wiki}
\end{subfigure}
\caption{Wiki: Precision of top $k$ candidates vs. $k$ for different evidence detection models.}
\label{fig:Wiki_test}
\end{figure}

\subsubsection{Wiki}
Following the high performance of the end-to-end system on VLD, it is interesting to examine how the results are affected by moving to the smaller and more homogeneous corpus of Wiki. For this purpose we tested the models trained using VLD, on Wiki test-set. In addition we trained new models on Wiki training-set and tested them on 
the Wiki 
test-set.
Figure \ref{fig:Wiki_test} shows the average precision of our models vs. the number of top candidates. 
Although the precision values of the leading model are still high, they are significantly lower than those obtained in VLD. 
One possible reason is the difference in domains, that is, that the distributions of sentences retrieved from VLD and those retrieved from Wiki are somewhat different. Another reason may be that the dramatically smaller number of sentences in Wiki is likely to result in lower scores among the top ranked candidates\footnote{Let $X$ be the number of sentences; the top $k$ candidates correspond to the top $k/X$ percentile of scores. For $Y<X$, the percentile of the top $k$ candidates becomes $k/Y> k/X$, which corresponds to lower scores.}. The lower the score the less 
likely it is that a sentence is
evidence. Indeed, the Wiki scores over the top k predictions are significantly lower than the VLD scores (t-test p-value=$3.19e^{-9}$ for $k=20$). Despite the performance differences when moving from VLD to Wiki, the hierarchy between the different models is the same. Moreover, although the test sentences are from Wiki, the models which are trained on Wiki are inferior to those trained on VLD. This can be attributed to the much larger training data of VLD (154K compared to 22K in Wiki), which compensates for the potential difference between the train and test distributions. 
\subsection{BlendNet Benchmark}
The \BAMaskS network used here has the same architecture as that in \citet{shnarch-etal-2018-will}. It is therefore interesting to evaluate how the differences between their work and ours - greatly expanding the training data, considering the motion explicitly and using the BERT architecture - are reflected over the benchmark based on their data. 
Table \ref{tab:acc} describes the impact of these differences. Training BA$_{MaskS}$ on the VLD improves accuracy from 74\% to 78\%. 
In other words, even though BlendNet was trained on Wikipedia sentences, and the benchmark is also composed of Wikipedia sentences, the larger size of the VLD more than compensates for the change in domain.
Shifting to the BERT architecture improves accuracy to 81\%, even without adding the motion as part of the input. Finally, the best accuracy is attained when combining all three modification, for a total of 
nearly $14\%$ 
improvement over the 
best results reported by \citet{shnarch-etal-2018-will}.  

\begin{table}[t] 
\begin{center}
\begin{tabular}{|c|c|c|}  

\hline \bf Arch. &\bf trained on & \bf Accuracy\\ 
\hline
\BAMaskS & \cite{shnarch-etal-2018-will} & 0.74 \\
\hline
\BAMaskS & VLD & 0.78 \\
\hline
\BAMaskSM & VLD & 0.77 \\
\hline
\BertMaskS & VLD & 0.81 \\
\hline
\BertMaskSM & VLD & 0.82 \\
\hline
\BertSM & VLD & 0.84 \\
\hline

\end{tabular}
\end{center}
\caption{Accuracy of sentence classification over BlendNet.}
\label{tab:acc}
\end{table}

\subsection{UKP-TUDA Benchmark}
The {\em UKP-TUDA} benchmark is an interesting case study for our networks, since it is labeled for whether or not a sentence is argumentative (defined as having a clear stance w.r.t. the topic), rather than whether or not it is evidence. In particular, argumentative texts which are not {\em Study Evidence} or {\em Expert Evidence} are labeled as positive examples, but are expected to receive a low score by \BertSM. In other words, when naively used as a classifier, \BertSM is expected to accept argumentative sentences which are evidence of the appropriate type, but reject argumentative sentences which are not.

Indeed, a threshold of $0.5$ leads to a precision score
of $0.88$ and a recall score of $0.16$. To verify that the predicted positives indeed tend to be {\em evidence} and that the low recall is due to other types of argumentative sentences being rejected, we manually annotated $20$ {\em argumentative} sentences with a score above the threshold, and $20$ with a score below it (chosen uniformly at random). As expected, in the first set, 14 of the sentences are {\em Study Evidence} or {\em Expert Evidence}, while in the second set only 2 are so.

Importantly, \BertSM was trained on sentences which correspond to 
the retrieval queries, and are therefore quite different from the sentences in the {\em UKP-TUDA} benchmark; In particular, it is not clear what fraction of the negative training examples are non-argumentative sentences. Hence, it is 
somewhat unclear 
how such sentences will be scored. To evaluate this, we considered all sentences with a classification score below the threshold. Among those which are argumentative, the average score is $7.3\cdot10^{-2}$, while among those which are not argumentative it is $1.5\cdot10^{-2}$ (t-test p-value $< 10^{-76}$), suggesting that argumentative sentences tend to be assigned a higher score.

\begin{figure}[ht!]
\centering
\includegraphics[width=.95\columnwidth]{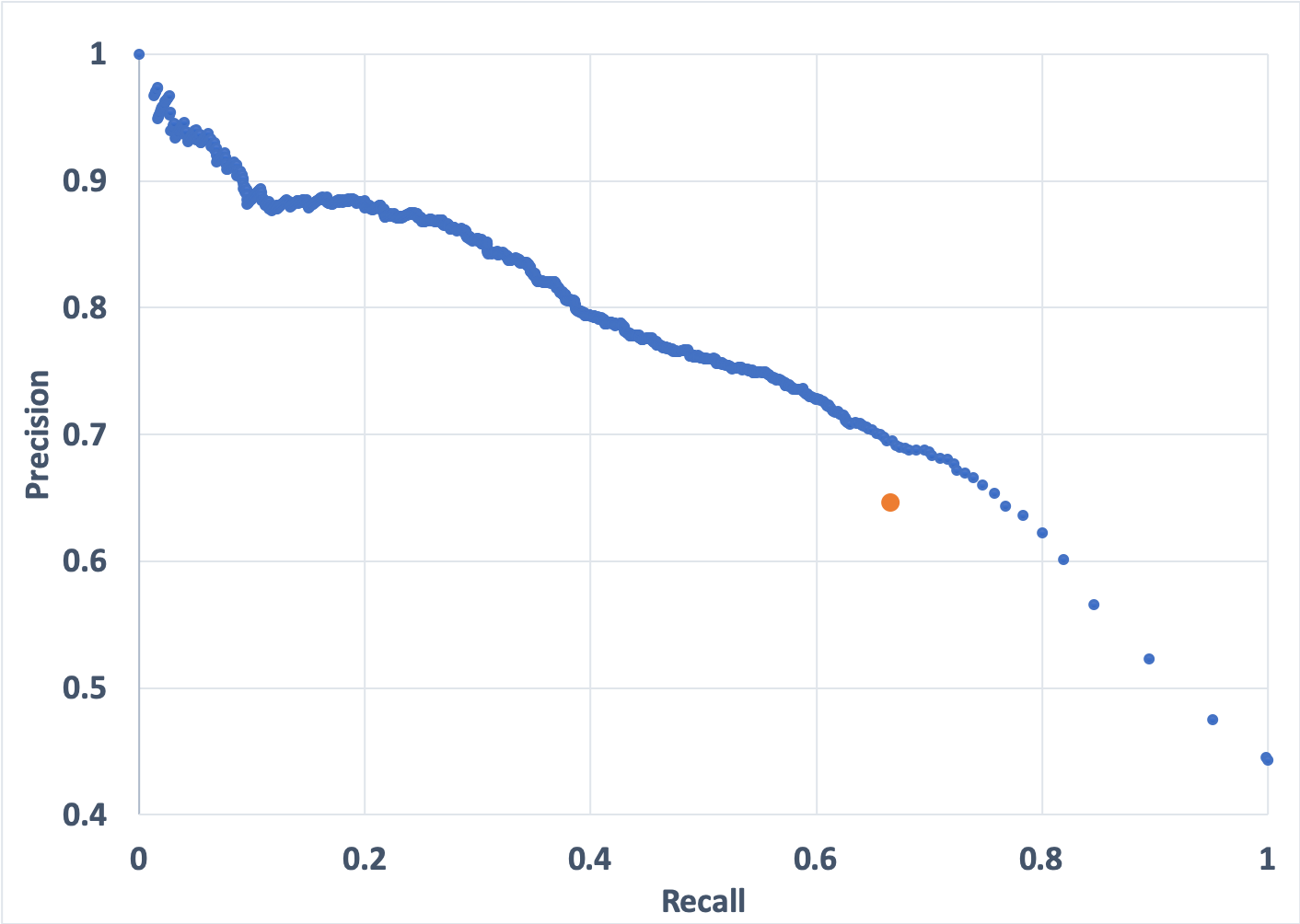}
\caption{Precision-Recall curve of \BertSM over the {\em UKP-TUDA} benchmark. The orange mark denotes the best result reported by \citet{stab2018cross}.
\label{fig:stab}}
\end{figure}

A different way to examine this phenomenon is by considering the precision-recall curve over this benchmark. Figure \ref{fig:stab} shows that even though the network was trained to discern {\em evidence} sentences, it also learned to prefer argumentative sentences over non-argumentative ones. The curve-points toward its right end correspond to precision-recall trade-offs similar to those reported in \citet{stab2018cross} for classifiers trained directly for this task. For example, a threshold score of $0.002$ corresponds to precision of $0.66$ and recall of $0.75$ (F1-score=$0.7$) on this curve. The preferred network in \citet{stab2018cross} achieves an average precision of $0.65$, $0.67$ average recall and $0.67$ average F1-score (see Figure \ref{fig:stab}).

Taken together these results suggest that even though \BertSM was trained on rather different data, its ranking of the {\em UKP-TUDA} tends to be {\em Study Evidence} or {\em Expert Evidence} first, then argumentative sentences not of these types and last but not least - the non-argumentative sentences. 

\begin{figure*}[!t]
\newcolumntype{P}[1]{>{\raggedright\arraybackslash}p{#1}}
\begin{center}
\includegraphics[width=.95\textwidth]{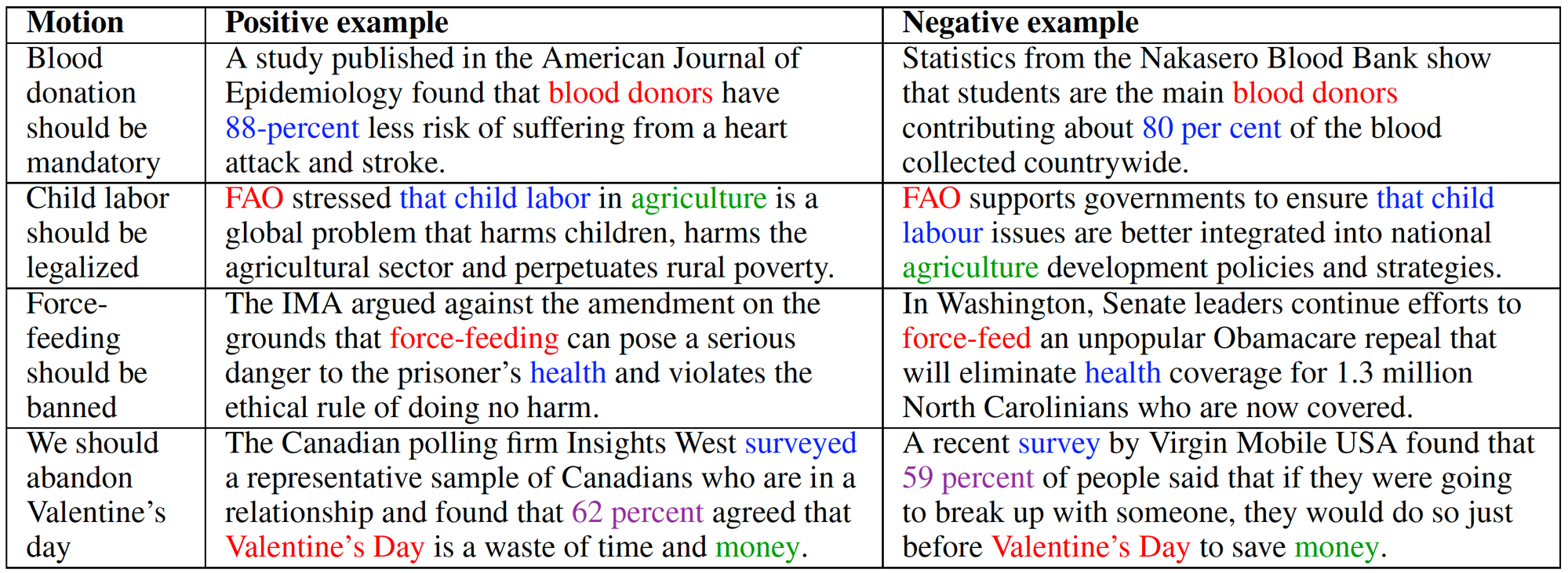}
\end{center}
\caption{Example of sentences containing similar terms, of which only one is a relevant evidence. Similar terms are in the same color.}
\label{tab:examples}
\end{figure*}

Finally, to appreciate that the task of retrieving relevant evidence can be quite nuanced and challenging, 
the table in Figure \ref{tab:examples} lists several motions, each with a pair of retrieved sentences. In each case, the pair of sentences contain similar terms, but nonetheless one is a relevant evidence, and the other is not. All these examples were successfully discerned by the suggested model.

\section{Discussion and Conclusion}
We presented a 
first end-to-end "working solution" 
for the argument retrieval task, showing that it attains remarkably 
high precision on a wide range of topics. 
Moreover, in spite of being trained on data of a somewhat different nature, our best model attained results comparable to previous state-of-the-art in one benchmark data \cite{stab2018cross}, and significantly outperformed state-of-the-art on a second benchmark data \cite{shnarch-etal-2018-will}.
A key element of our approach 
is the \textit{Retrospective Labeling} 
paradigm, a precision-oriented Active Learning, which is targeted to cope with skewed label distribution. This strategy is of general applicability, especially in retrieval tasks where precision is the common evaluation metric and positive examples are scarce.

We have suggested the SL approach as an alternative to the document-based one, but it can actually complement it. One can envision a hybrid solution that may enjoy the best of both worlds. 
One approach could be to simply combine sentences retrieved by both methods. Alternatively, one could start with SL retrieval, and then leverage the SL predictions to identify relevant documents or passages. Specifically, one could run the SL based system suggested here, and then expand each of the top-ranking sentences into the containing  
paragraph or document. Presumably, such a method would enjoy both the wide variety of contexts in which SL-based arguments are found, while alleviating the restrictions that retrieved sentences conform to one of the SL-queries. 

Another concession that was made here in the interest of precision is limiting the scope of retrieval to sentences conforming to the queries' patterns. It is interesting to try and expand this, in particular to sentences which do not mention the {\em topic} explicitly. One approach could be to try and solve the co-reference problem during indexing, thus being able to retrieve sentences in which the {\em topic} is only referenced. 
Another approach would be to first expand the topic, and then do argument-retrieval for the expanded set of topics \cite{bar2019surrogacy}.

Our work focused on the retrieval of specific types of Evidence, but the same approach can be applied to other types of arguments. In particular, 
we used the same approach 
to retrieve {\em Claims} with high precision; the details of these analyses are omitted due to lack of space.

Moreover, argument-retrieval can be defined at different levels of granularity - e.g., retrieving arguments of any type, retrieving only {\em Evidence}, retrieving only {\em Study Evidence}, or retrieving only certain types of {\em Study Evidence} such as polls. Arguments of different types share many commonalities due to their argumentative nature, as shown by employing the model developed here on the data of \citet{stab2018cross}. Accordingly, Transfer Learning techniques can probably exploit the large annotated dataset herein for the task of retrieving arguments of other types. 

Similarly, it is interesting to explore how Domain Adaptation techniques can leverage labels from an annotated corpus such as 
those 
presented here to novel ones.
Namely, while we have shown that obtaining high-quality, balanced labeled data from a massive corpus is plausible, it does nonetheless require a considerable annotation effort. Hopefully, when presented with a corpus from a new domain, the models developed here can be adapted to effectively retrieve arguments from that corpus as well.

\fontsize{9.0pt}{10.0pt} \selectfont
\bibliography{alt_main}
\bibliographystyle{aaai}
\appendix

\end{document}